\documentclass[runningheads]{llncs}
\usepackage{graphicx}
\usepackage{booktabs}
\usepackage{amssymb}
\usepackage{multirow}
\usepackage{color}
\usepackage{amsmath}
\usepackage{algorithmic}
\usepackage[misc]{ifsym}
\usepackage{booktabs}
\usepackage[colorlinks]{hyperref}
\usepackage[symbol]{footmisc}
\begin{document}
\title{Computer-aided Tuberculosis Diagnosis with Attribute Reasoning Assistance}
%
%

\author{Chengwei Pan\inst{1} \and Gangming Zhao\inst{2,3}\thanks{C. Pan and G. Zhao contributed equally to this work.} \and Junjie Fang\inst{4,5} \and Baolian Qi\inst{5} \and Jiaheng Liu\inst{1} \and \\ Chaowei Fang\inst{6} \and Dingwen Zhang\inst{7} \and Jinpeng Li\inst{5}\thanks{Corresponding author. Email: lijinpeng@ucas.ac.cn}$^{\left(\textrm{\Letter}\right)}$ \and Yizhou Yu\inst{2, 3}
}
%
\authorrunning{C. Pan \it{et al.}}
%
\institute{\textsuperscript{1}Institute of Artificial Intelligence, Beihang University, Beijing, China\\
\textsuperscript{2}The University of Hong Kong, Hong Kong, China \\
\textsuperscript{3}Deepwise AI Lab, Beijing, China \\
\textsuperscript{4}Hwa Mei Hospital, UCAS \\
\textsuperscript{5}Ningbo Institute of Life and Health Industry, UCAS\\
\textsuperscript{6}School of Artificial Intelligence, Xidian University, Xi’an, China\\
\textsuperscript{7}School of Automation, Northwestern Polytechnical University, Xi'an, China}

\maketitle              
\begin{abstract}
 Although deep learning algorithms have been intensively developed for computer-aided tuberculosis diagnosis (CTD), they mainly depend on carefully annotated datasets, leading to much time and resource consumption. Weakly supervised learning (WSL), which leverages coarse-grained labels to accomplish fine-grained tasks, has the potential to solve this problem. In this paper, we first propose a new large-scale tuberculosis (TB) chest X-ray dataset, namely tuberculosis chest X-ray attribute dataset (TBX-Att),
 and then establish an attribute-assisted weakly supervised framework to classify and localize TB by leveraging the attribute information to overcome the insufficiency of supervision in WSL scenarios. Specifically, first, the TBX-Att dataset contains 2000 X-ray images with seven kinds of attributes for TB relational reasoning, which are annotated by experienced radiologists. It also includes the public TBX11K dataset with 11200 X-ray images to facilitate weakly supervised detection. Second, we exploit a multi-scale feature interaction model for TB area classification and detection with attribute relational reasoning. The proposed model is evaluated on the TBX-Att dataset and will serve as a solid baseline for future research. The 
code and data will be available at ~\url{https://github.com/GangmingZhao/tb-attribute-weak-localization}.
\end{abstract}

\section{Introduction}
Tuberculosis (TB) is one of the most serious thoracic diseases associated with a high death rate. The early diagnosis and treatment is very important to confront the threat of TB. Computer-aided tuberculosis diagnosis (CTD) has been widely investigated to assist radiologists in diagnosing TB. Although deep learning algorithms have achieved stupendous success in automatic disease classification~\cite{wang2017chestx,aviles2019graphx} and localization~\cite{cai2018iterative,li2018thoracic,liu2019align} for chest X-rays, their performance on CTD remains a barrier to clinical application. One obstacle lies in the absence of high quality annotated datasets and the scarcity of clinical features (attributes). In practice, such attributes have been validated to be helpful to the performance and interpretability. 

In addition, the main challenges of diagnosing TB with chest X-ray images include the low visual contrast between lesions and other regions, and distortions induced by other overlapping tissues. Sometimes it is difficult for radiologists to recognize obscure diseases.
 For CTD, previous methods mainly concentrated on disease classification~\cite{Qin2018,brestel2018radbot}
, and several recent works have taken a step forward to detect disease regions under weak/limited supervisions. They can be grouped into two main categories: the first category ~\cite{wang2017chestx}
resorts to convolutional neural networks (CNN) trained on the classification task and output disease localization results through calculating the category activation maps ~\cite{zhou2016learning}; the second category ~\cite{li2018thoracic,liu2019align} uses the multiple instance learning to directly yield categorical probability maps that can be easily transformed to lesion positions. However, the performance of these methods is still far from clinical usage. Effective WSL algorithms remain the boundary to explore.

The mainstream pipeline of object detection is screening out potential proposals followed by proposal classification~\cite{ren2015faster}. By stacking piles of convolutional layers, CNNs are very advantageous at extracting surrounding contextual information, however, distant relationships are still hard to capture by convolutions with small kernels. On the other hand, radiologists use attribute information for diagnosis, whereas the pure data-driven CNNs can not take this prior knowledge into account, leading to disconnections to the clinical practice.

The main contributions of this paper can be summarized as follows.
\begin{itemize}
    \item We present a large-scale TBX-Att dataset for attribute-assisted X-ray diagnosis for TB. To the best of our knowledge, this offers the first playground to leverage the attribute information to help models detect TB areas in a weakly-supervised manner.
    \item We propose an effective method to fuse the attribute information and TB information. The multi-scale attribute features are extracted for each TB proposal under the guidance of a relational reasoning module and are further refined with the attribute-guided knowledge. A novel feature fusion module is devised to enhance the TB representation with its effective attribute prompt.
    \item The proposed method improves object detection baselines~\cite{ren2015faster,lin2017feature} by large margins on TBX-Att, leading to a solid benchmark for weakly supervised TB detection.
\end{itemize}

\section{Related Work}
\subsection{Object Detection}
Object detection is a widely-studied topic in both natural and medical images. It aims at localizing object instances of interest such as faces, pedestrians and disease lesions. The most famous kind of deep learning approaches for object detection is the R-CNN~\cite{girshick2014rich} family. The primitive R-CNN extracts proposals through selective search~\cite{uijlings2013selective}, and then predicts object bounding boxes and  categories from convolution features of these proposals. Fast R-CNN~\cite{girshick2015fast} adopts a shared backbone network to extract proposal features via RoI pooling. Faster R-CNN~\cite{ren2015faster} automatically produces object proposals from top-level features with the help of pre-defined anchors. The above methods accomplish the detection procedure through two stages, including object proposal extraction, object recognition and localization. In~\cite{lin2017feature}, the feature pyramid network is exploited to further improve the detection performance of Fast R-CNN and Faster R-CNN with the help of multi-scale feature maps.
The other pipeline for object detection implements object localization and identification in single stage through simultaneous bounding box regression and object classification, such as YOLO~\cite{redmon2016you} and SSD~\cite{liu2016ssd}. The RetinaNet~\cite{lin2017focal} is also built upon the feature pyramid network, and uses dense box predictions during the training stage. The focal loss is proposed to cope with the class imbalance problem.
The detection task has also attracted a large amount of research interest in medical images, such as lesion detection in CT scans~\cite{yan2018deeplesion} and cell detection in malaria images~\cite{hung2017applying}. This paper targets at detecting diseases in chest X-ray images. Practically, we propose a \textit{TBX-Att dataset} to exploit attribute information to enhance TB feature representations of disease proposals.

\subsection{Disease Diagnosis in Chest X-ray Images}
Accurately recognizing and localizing diseases in chest X-Ray images is very challenging because of low textural contrast, large anatomic variation across patients, and organ overlapping. Previous works in this field mainly focus on disease classification~\cite{noor2014texture,wang2017chestx,aviles2019graphx,9627588}. Recently, the authors in~\cite{chouhan2020novel} propose to transfer deep models pretrained on the ImageNet dataset~\cite{deng2009imagenet} for recognizing pneumonia in chest X-ray images. In~\cite{sahlol2020novel}, the artificial ecosystem-based optimization algorithm is used to select the most relevant features for tuberculosis recognition.
Based on the category activation map~\cite{zhou2016learning} which can be estimated with a disease recognition network, researchers attempt to localize disease in a weakly supervised manner~\cite{wang2017chestx}.
In~\cite{zhang2020thoracic}, the triplet loss is used to facilitate the training of the disease classification model, and better performance is observed in class activation maps (CAM) estimated by the trained model.
In~\cite{li2018thoracic,liu2019align}, multiple instance learning is employed to solve the disease localization problem.
In~\cite{zhou2018weakly}, a novel weakly supervised disease detection model is devised on the basis of the DenseNet~\cite{huang2017densely}. Two pooling layers including a class-wise pooling layer and a spatial pooling layer are used to transform 2-dimensional class attention maps into the final prediction scores. The performance of these methods is still far from practical usage in automatic diagnosis systems. In~\cite{liu2020rethinking}, a new benchmark is proposed to identify and search potential TB diseases, however, it lacks attribute information that has been proven to pose a great effect on the practice diagnosis of medical experts.

\begin{figure*}[htb]
\centering
\includegraphics[width=1\textwidth]{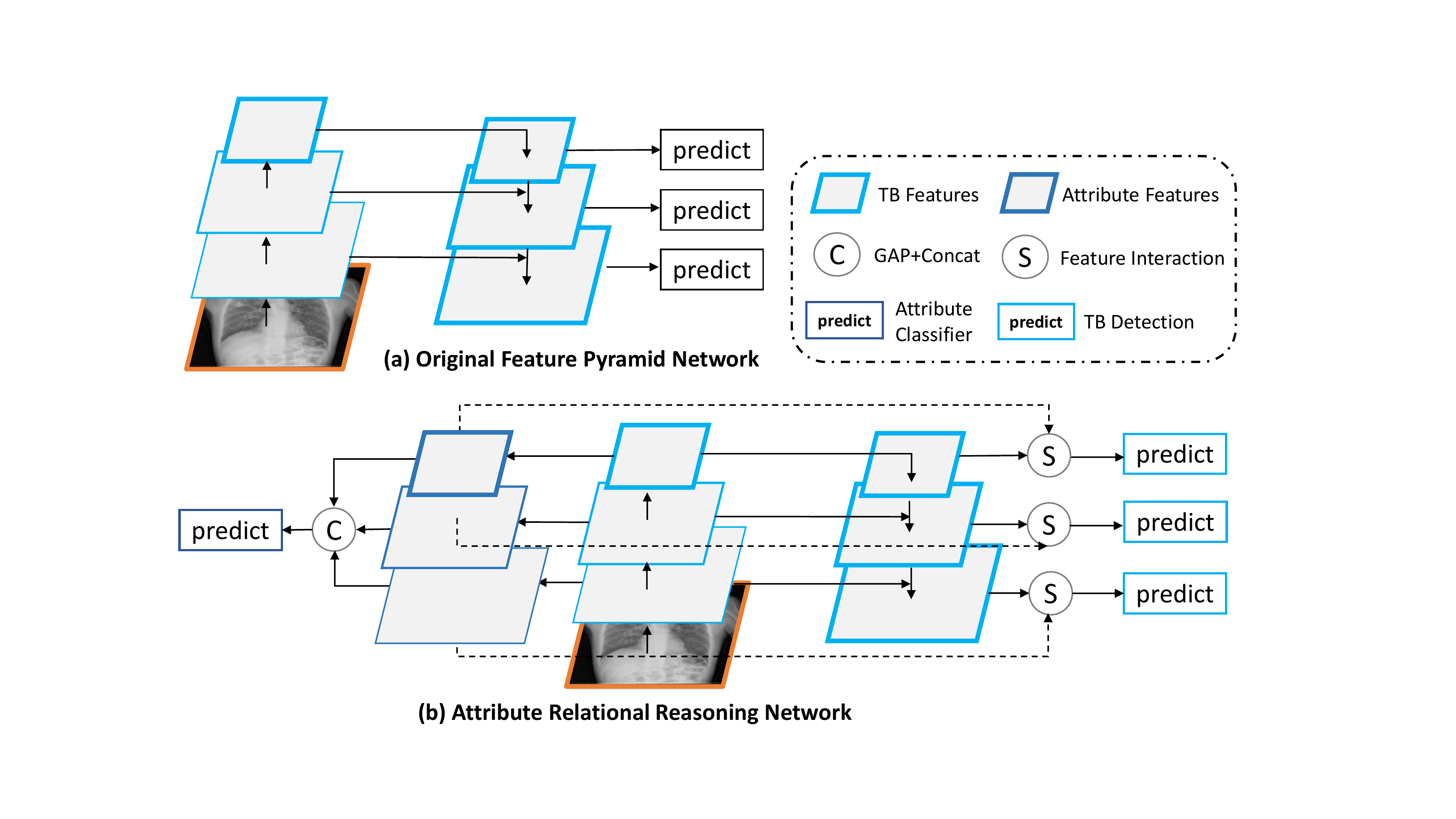}
\caption{\label{framework} Illustration of the proposed Attribute Relational Reasoning Network.}
\end{figure*}

\section{Methods}\label{sec:met}
\subsection{Overview}
The method proposed in this paper mainly addresses the weakly supervised tuberculosis detection from chest X-rays. The pipeline of our proposed method is highlighted in Fig.~\ref{framework}, which include three parts: (1) the feature pyramid network is utilized for extracting multi-scale tuberculosis feature maps; (2) the attribute classifier, which employs the attribute supervised information to generate multi-scale attribute feature maps; (3) the feature interaction module takes the attribute feature map to establish the attribute prompt to obtain a more representative feature representation of tuberculosis. Compared with the original Feature Pyramid Network used for object detection, another branch is added to generate the corresponding multi-scale features for attribute classification, which can prompt a guidance to extract representative features for detection. 

\subsection{Attribute Feature Representation}

\begin{figure}[htb]
\centering
\includegraphics[width=0.8\linewidth]{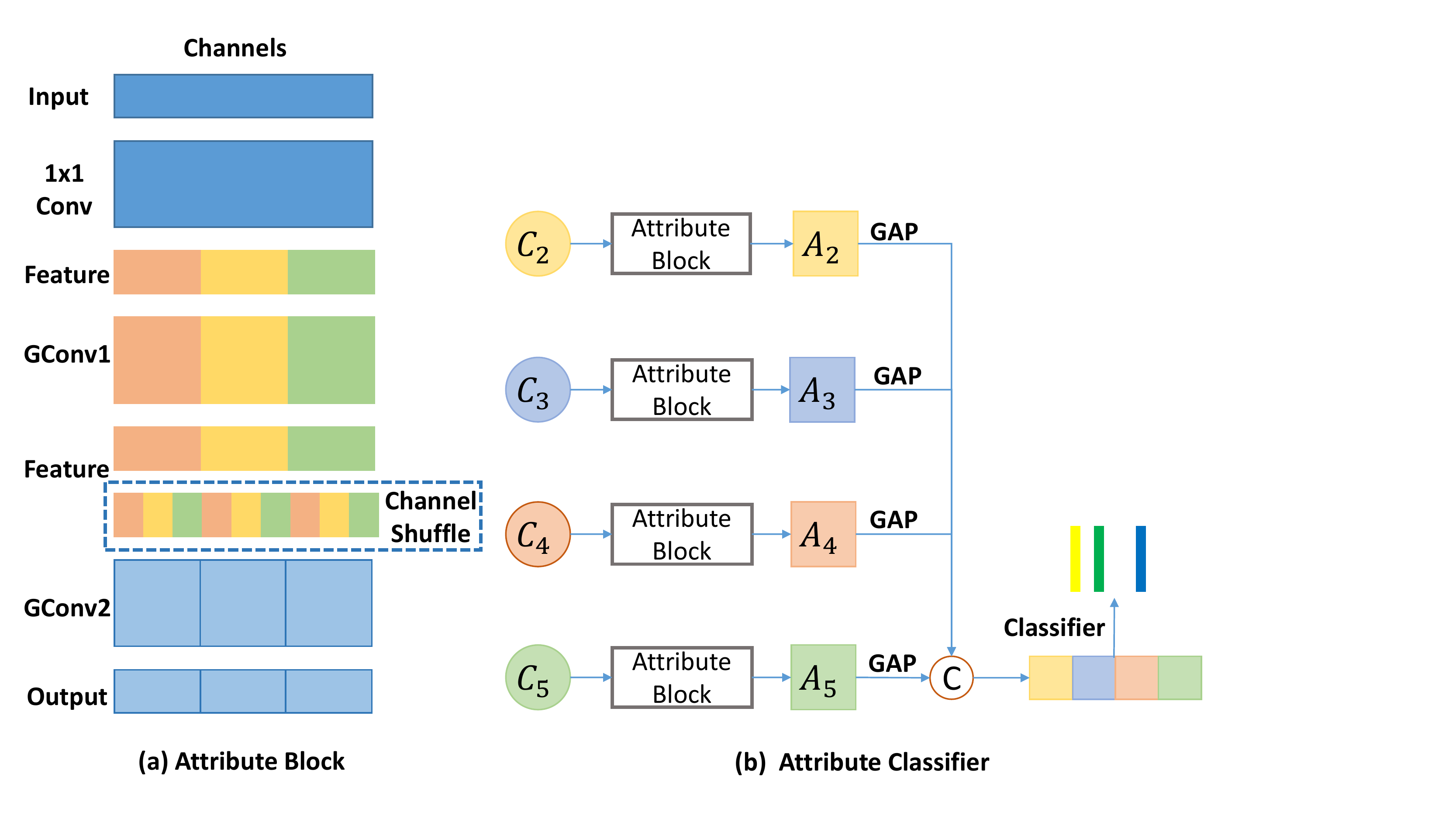}
\caption{\label{attribute} Attribute feature representation. (a) shows the process of attribute feature extraction. (b) shows the fusion of multi-scale attribute features and the classification.}
\end{figure}

The detection branch is designed to utilize multi-scale features to generate foreground detection result. The backbone network used for abstracting feature representation always generate multi-scale features. Specifically, for ResNets we use the feature activation output by each stage's last residual block. We denote the output of these blocks as $\{C_2, C_3, C_4, C_5\}$, which have strides of $\{4,8,16,32\}$ pixels with respect to the input image. Correspondingly, multi-scale attribute features as shown in Fig.~\ref{attribute} are constructed in our method for both the task of classification and feature interaction described in the next section.

Given the input $C_i$ from the i-th stage's last residual block, we simply attach a $1\times1$ convolution layer to produce a feature map having $C_a \times N_a$ channels, where $N_a$ represents the number of kinds of attributes and $C_a$ represents the channel dimension of each attribute feature map. Then two group convolution layers are used to generate the distinguishable feature maps for each attribute individually. For each group convolution layer, we set the number of groups as $N_a$ and expect each group to represent a kind of attribute. For better taking advantage of information flowing across attributes' feature maps, channel shuffle operation is added in the last group convolution layer. The specially designed Attribute Block used for extracting attribute features is shown in Fig. \ref{attribute}(a). 

After obtaining the features of each scale($A_2$ to $A_5$ in Fig. \ref{attribute}(b), the Global Average Pooling operation and feature fusion by concatenation are used to get the final feature vectors used for attribute classification.

\subsection{Feature Interaction}

\textbf{Multi-head Cross-Attention} Given two feature maps $\textbf{X}$, $\textbf{Y}\in R^{C \times H \times W}$, where $H$,$W$ are the spatial height, width and $C$ is the number of channels. One $1 \times 1$ convolution is used to project $\textbf{X}$ to query embedding $\textbf{Q}_i$, another two $1 \times 1$ convolutions are adopt similarly to project $\textbf{Y}$ to key embedding $\textbf{K}_i$ and value embedding $\textbf{V}_i$, where $i$ represents the i-th head. The dimension of embedding in each head is $d$, and the number of heads is $N=C/d$. The $\textbf{Q}_i$, $\textbf{K}_i$, $\textbf{V}_i$ are then flattened and transposed into sequences with the size of $n \times d$, where $n=H \times W$. The output of the cross-attention (CA) in i-th head is a scaled dot-product:

\begin{equation}
    CA_i(\textbf{Q}_i, \textbf{K}_i, \textbf{V}_i) = 
    SoftMax(\frac{\textbf{Q}_i\textbf{K}_i^T}{\sqrt{d}})\textbf{V}_i.
\end{equation}
To reduce the large computational complexity of CA on high resolution feature maps, $\textbf{K}_i$ and $\textbf{V}_i$ are down-sampling by the ratio of $s$, resulting smaller feature maps with the size of $d \times \frac{H}{s} \times \frac{W}{s}$. In our implementation, $s$ is set to 16 in the first stage, and gradually reduce by 2 times. Finally, in a $N$-head attention situation, the output after multi-head cross-attention (MCA) is calculated as follows: 
\begin{equation}
    MCA(\textbf{X}, \textbf{Y}) = \Phi(Concat(CA_1, CA_2, ..., CA_N))
\end{equation}
where $\Phi(\cdot)$ is a liner projection layer that weights and aggregates the feature representation of all attention heads.

\subsubsection{Attr-Attr Attention ($A^2$-Attn)}
\begin{figure}[htb]
\centering
\includegraphics[width=1.0\linewidth]{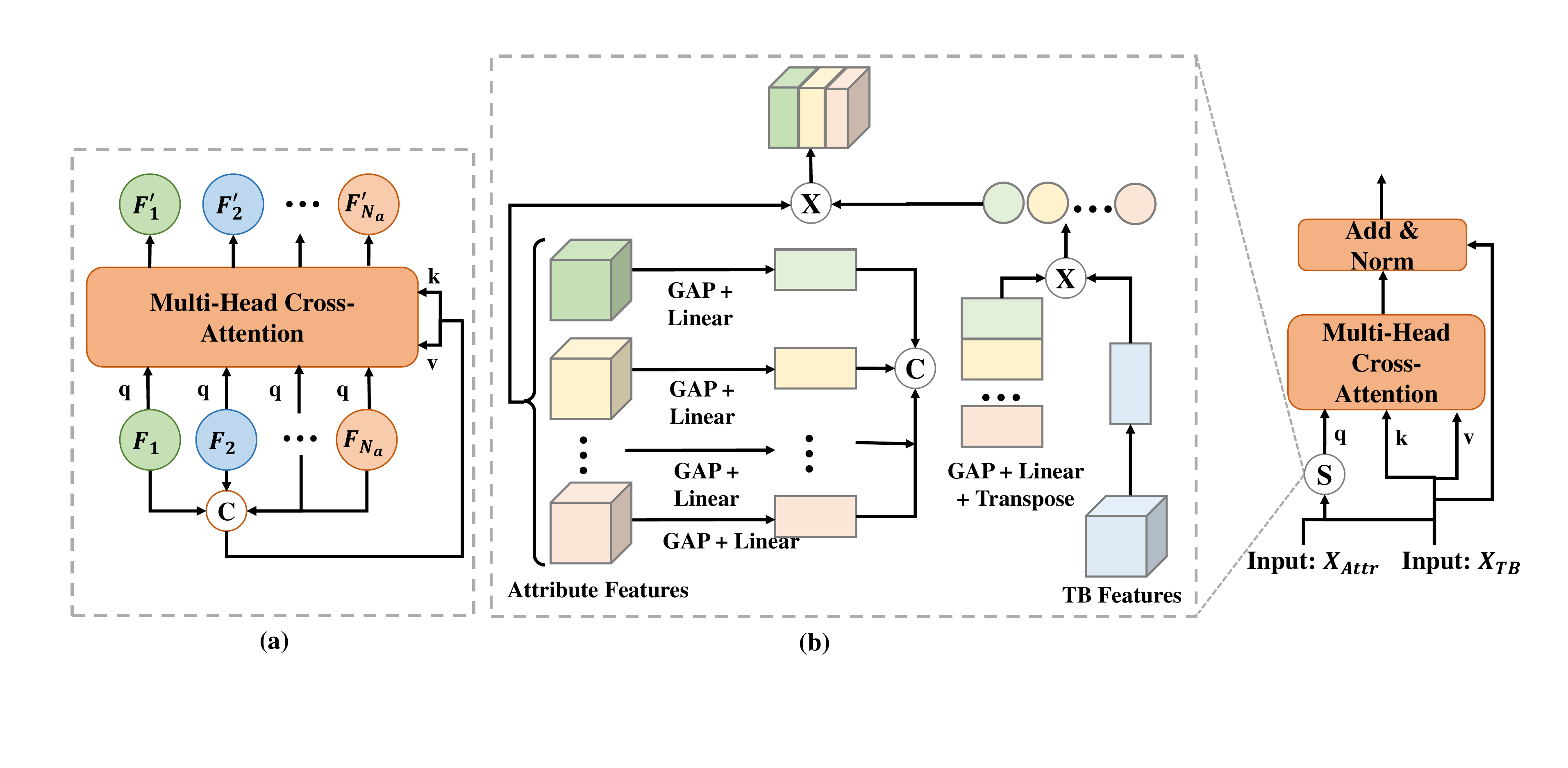}
\caption{\label{self_attn} Cross-attention. (a) shows the multi-head cross attention among different kinds of attribute features. (b) shows cross attention between attribute features and TB features.}
\end{figure}
Given the input $\textbf{X} = \{\textbf{F}_1,\textbf{F}_2,...,\textbf{F}_{N_a}\}$ from the i-th scale of attribute feature maps, attention modules are used to reproduce more representative features by implicitly mining the relationship among $N_a$ kinds of attributes. To build the attention module, we firstly concatenate $\textbf{F}_i$ channel-wisely, and then linearly projected to get $\textbf{Y}$, which has the same channel dimension of $\textbf{F}_i$. As shown in Fig. \ref{self_attn}(a), the attention module takes $\{\textbf{F}_i, i=1,...,N_a\}$ as queries and uses $\textbf{Y}$ to generate key and value embeddings, so the finally aggregated feature $\textbf{F}_{i}^{'}$ can be obtained by $\textbf{F}_{i}^{'} = MCA(\textbf{F}_i, \textbf{Y}$). The operation of $A^2$-Attn can be applied to attribute features of each scale ($A_2$ to $A_5$ in Fig. \ref{attribute}(b)) to get more representative features. 

\subsubsection{Attr-TB Attention (AT-Attn)}
Given attribute features $\{\textbf{F}_i, i=1,...,N_a\}$ and TB feature $\textbf{F}_{tb}$, we expect to design a attribute and TB attention module (AT-Attn) to enhance the TB representation with attribute prompt. We firstly reaggregate attribute features to get \textbf{X} by taking advantage of the similarity between attribute features and TB feature. Specifically, 
GAP operation followed by a linear projection is performed to get attribute vectors $\{\textbf{A}_i, i=1,...,N_a\}$ and TB vector $\textbf{B}$. The similarity score $s_i$ is calculated by $\textbf{A}_i \cdot \textbf{B}$, and then \textbf{X} can be obtained by $\textbf{X} = \sum s_i\textbf{F}_i$. As shown in in Fig. \ref{self_attn}(b), the attention module takes $\textbf{X}$ as query and use $\textbf{F}_{tb}$ to generate key and value embeddings. Finally, the output of the AT-attn is obtained as follows:
\begin{equation}
    \textbf{F}_{tb}^{'} = \textbf{F}_{tb}+Norm(MCA(\sum s_i\textbf{F}_i, \textbf{F}_{tb}))
\end{equation}

\subsection{Loss Function}
During the training, the classification branch and the detection branch are jointly trained, which can be seen as a type of multi-tasking learning. $Loss_{det}$ is used to optimize the detection branch, and $Loss_{cls}$ is performed to train the classification branch. For $Loss_{cls}$, we apply the most commonly used binary cross-entropy loss. For $Loss_{det}$, the same loss function is used as in \cite{ren2015faster}, which contains an anchor category prediction item and a bounding box regression item. The final loss function is the sum of $Loss_{det}$ and $Loss_{cls}$, and $\lambda$ is used for balancing the influence of two loss terms.

\begin{equation}
    Loss = Loss_{det} + \lambda Loss_{cls}
\end{equation}

\subsection{Experimental Analysis}

\subsubsection{Dataset}
The collection of attribute-based TB X-rays faces many difficulties~\cite{liu2020rethinking}. The attribute information is so hard to obtain due to its unclear pathological structure. To reduce these problems, we cooperate with one of the most important hospitals in China to collect X-rays. We also combine the TBX11K dataset~\cite{liu2020rethinking} to construct a totally new attribute-based TB dataset. The TBX11K dataset consists of 11200 X-rays, including 5000 healthy cases, 5000 sick but non-TB cases, and 1200 cases with manifestations of TB. The dataset includes 8 types of sign: Fibrotic Streaks, Pulmonary Consolidation, Diffuse Nodules, Pulmonary Cavitation, Atelectasis, Multiple Nodules, Pleural Effusion, and Pulmonary Tuberculosis. Note that all X-ray images are resized and have the resolution of about 512 $\times$ 512. Two senior radiologists participated in the annotation independently following a list of attributes. If an attribute is present, it is annotated as 1 and otherwise 0, resulting in a sparse vector of attributes. The sample was retained only if the two radiologists labeled the same. The attribute dataset consists of a toal of 2000 X-ray images, including a training set of 1700 images and a validation set of 300 images. For the combination with the existing TBX11K dataset, annotations from the attribute dataset are used to train the attribute classification branch, while the TBX11K dataset consists of TB localization information which provides the supervision for training the detection branch.

\subsubsection{Settings}
We adopt Faster RCNN as baseline, in addition, a single scale feature interaction is firstly used to explore the influence of cross attribute attention. Then a multi-scale feature interaction way is adopt to further enhance the attention function. For all experiments, each model is independently trained five times with randomly initialized weights and correspondingly 5 validations are performed for each model. The overall performance of a model is assessed with several commonly used metrics, including the mean and standard deviation of accuracy, F-score and mAP. All models are trained for 60 epoches from scratch using PyTorch~\cite{paszke2019pytorch} on NVIDIA Titan X pascal GPUs while Adam~\cite{kingma2014adam} being the optimizer with the initial learning rate set to 1e-3, which is reduced by a factor of 10 after every 20 epochs. The weight decay is set to 1e-4. We adopt ResNet-50 as the backbone model. The batch size is 8 on a single GPU. Different from ~\cite{liu2020rethinking}, we train the two branches simultaneously rather than dividing it into two stages.


\begin{table*}[t]\normalsize
	\caption{Performance Comparison of Tuberculosis Diagnosis Models on the TB-Xatt Dataset.}
\centering
\scalebox{0.65}{%
	\centering
\begin{tabular}{c|c|c|c|c|c|c|c}
  \hline
\toprule[2pt]
\multirow{2}{*}{Methods}   & \multirow{2}{*}{Detector}  & \multicolumn{3}{c|}{Major Component} & \multicolumn{3}{c}{Results(\%) (mean$\pm$standard deviation)} \\
 \cline{3-8}
  & &GroupConv & $A^{2}$-Attn & AT-Attn & F-score & Accuracy & mAP  \\
  \hline \hline
Baseline & Two-stage Model & $\boxtimes$ &$\boxtimes$ &  $\boxtimes$  & 29.24$\pm$0.76 & 88.08$\pm$0.12 & 17.10$\pm$0.11\\
\midrule
\multirow{3}{*}{SingleScale} & \multirow{3}{*}{Two-stage Model} & $\boxtimes$ & $\checkmark$ &  $\checkmark$  & 33.70$\pm$0.63 & 92.08$\pm$0.71 & 18.20$\pm$0.17\\

& & $\checkmark$ &$\boxtimes$ &  $\checkmark$  & 32.70$\pm$0.72 & 91.08$\pm$0.12 & 17.20$\pm$0.16\\

& & $\checkmark$&  $\checkmark$  &$\boxtimes$  & 32.42$\pm$0.12 & 91.03$\pm$0.64 & 17.09$\pm$0.13\\

& & $\checkmark$&  $\checkmark$  &$\checkmark$  & 34.24$\pm$0.63 & 93.15$\pm$0.21 & 19.00$\pm$0.12\\

\midrule
\multirow{3}{*}{MultiScale} & \multirow{3}{*}{Two-stage Model} & $\boxtimes$ & $\checkmark$ &  $\checkmark$  & 33.13$\pm$0.14 & 94.01$\pm$0.21 & 18.79$\pm$0.32\\

& & $\checkmark$ &$\boxtimes$ &  $\checkmark$  & 34.12$\pm$0.23 & 91.82$\pm$0.33 & 17.92$\pm$0.42\\

& & $\checkmark$&  $\checkmark$  &$\boxtimes$  & 34.22$\pm$0.11 & 92.01$\pm$0.24 & 18.01$\pm$0.22\\

& & $\checkmark$&  $\checkmark$  &$\checkmark$  & \textbf{39.37$\pm$0.12} & \textbf{94.61$\pm$0.11} & \textbf{19.20$\pm$0.10}\\
\bottomrule[2pt]
	\end{tabular}}
     \label{table:chestsota}
\end{table*}

\subsubsection{Results} On our new proposed TBX-Att dataset for weakly tuberculosis detection, we compared our attribute relational reasoning with a two-stage faster RCNN model. We focus on the impacts of three carefully designed components on both attribute classification and TB detection in experiments. As shown in Table~\ref{table:chestsota}, when the three components are all used, the highest performance is obtained and exceeds baseline by a large margin. Our proposed method under multi-scale setting achieves an accuracy of 94.61\%, a F-score of 39.37\% and a mAP of 19.20\%, and outperforms the baseline by 10.13\%, 6.53\% and 2.1\% respectively. Any pairwise combination will reduce the performance, which can demonstrate the effectiveness of each component implicitly. Some interesting phenomenon that the feature interaction can gain improvements on both tasks (not just TB detection) is found in Table ~\ref{table:chestsota}. Moreover, the results verify that the two tasks are complementary, where the higher accuracy of attribute classification leads to higher TB detection performance. And the above findings also confirm our original intention, which takes advantage of TB attribute information to guide the effective extraction of features for TB detection.

\section{Conclusion}
In this paper, we present a TBX-Att dataset with attribute labels to expand the existing TBX11K dataset for weakly tuberculosis detection, considering radiologists usually use clinical features like attribute information for diagnosis. Moreover, a multi-scale feature interaction model is devised to enhance TB feature representations under the guidance of relational knowledge reasoning. We find out that with the attribute information assistance, TB classification and detection are more easily to achieve a better performance. We hope this dataset can further inspire not only researchers but also medical experts. 

\subsubsection{Acknowledgements}
This work is supported by the National Natural Science Foundation of China (Grant Nos. 62141605, 62106248, U21B2048).
\bibliographystyle{splncs04}
\bibliography{egbib}

\end{document}